\documentclass[final]{cvpr}

\usepackage{times}
\usepackage{epsfig}
\usepackage{graphicx}
\usepackage{amsmath}
\usepackage{amssymb}
\usepackage{threeparttable}
\usepackage{multirow}
\usepackage{bm}

\usepackage[pagebackref=true,breaklinks=true,colorlinks,bookmarks=false]{hyperref}

\def\eg{{\em e.g.}}
\def\ie{{\em i.e.}}
\def\etal{{\em et al.}}

\def\cvprPaperID{3908} 
\def\confYear{CVPR 2021}
\begin{document}

\title{Detection, Tracking, and Counting Meets Drones in Crowds: A Benchmark}

\author{Longyin Wen$^{1,\ast}$, Dawei Du$^{2,}$\thanks{Both authors contributed equally to this work.}, Pengfei Zhu$^{3,}$\thanks{Corresponding author. This work was supported by the National Key Research and Development Program of China under Grant 2018AAA0102402, the National Natural Science Foundation of China under Grants 61732011, 61876127, and 61925602, Natural Science Foundation of Tianjin under Grant 17JCZDJC30800, the Applied Basic Research Program of Qinghai (2019-ZJ-7017).}, Qinghua Hu$^{3}$, Qilong Wang$^3$, Liefeng Bo$^1$, Siwei Lyu$^2$\\
$^1$JD Finance America Corporation, Mountain View, CA, USA\\ 
$^2$University at Albany, State University of New York, Albany, NY, USA\\
$^3$Tianjin University, Tianjin, China\\
{\tt\small \url{https://github.com/VisDrone/DroneCrowd}}}
\maketitle
\begin{abstract}
To promote the developments of object detection, tracking and counting algorithms in drone-captured videos, we construct a benchmark with a new drone-captured large-scale dataset, named as DroneCrowd, formed by $112$ video clips with $33,600$ HD frames in various scenarios. Notably, we annotate $20,800$ people trajectories with $4.8$ million heads and several video-level attributes. Meanwhile, we design the Space-Time Neighbor-Aware Network (STNNet) as a strong baseline to solve object detection, tracking and counting jointly in dense crowds. STNNet is formed by the feature extraction module, followed by the density map estimation heads, and localization and association subnets. To exploit the context information of neighboring objects, we design the neighboring context loss to guide the association subnet training, which enforces consistent relative position of nearby objects in temporal domain. Extensive experiments on our DroneCrowd dataset demonstrate that STNNet performs favorably against the state-of-the-arts.
\end{abstract}

\section{Introduction}
Drones, or general unmanned aerial vehicles (UAVs), equipped with cameras have been fast deployed to a wide range of applications, such as video surveillance for crowd control \cite{DBLP:conf/cvpr/ZhouWT12} and public safety \cite{DBLP:journals/cm/MotlaghBT17}. In recent years, many massive stampedes have taken place around the world that claimed many victims, making the automatic density map estimation, counting and tracking in crowds on drones important tasks, which draw great attention from the computer vision community.

Despite significant progress, crowd counting and tracking algorithms still have room for improvement to deal with drone-captured videos due to various challenges, such as view point and scale variations, background clutter, and small scales. Developing and evaluating these algorithms for drones are impeded by the lack of publicly available large-scale benchmarks. Some recent efforts \cite{DBLP:conf/cvpr/ZhangLWY15,DBLP:conf/cvpr/ZhangZCGM16,DBLP:conf/eccv/IdreesTAZARS18,DBLP:conf/wacv/ZhangSC18,DBLP:conf/icmcs/FangZCGH19,DBLP:journals/corr/abs-2001-03360} have devoted to construct datasets for crowd counting. However, the majority of them focus on crowd counting with still images or inconsistent frames by surveillance cameras, due to difficulties in data collection and annotation for drone-based crowd counting and tracking.

To fill this gap, we collect a large-scale drone-based dataset for density map estimation, crowd localization and tracking. Our DroneCrowd dataset consists of $112$ video clips formed by total $33,600$ frames, captured by various drone-mounted cameras, in $70$ different scenarios across $4$ different cities in China (\ie, Tianjin, Guangzhou, Daqing, and Hong Kong). These video clips are annotated with more than $4.8$ million head annotations and several video-level attributes. To the best of our knowledge, this is the largest and most thoroughly annotated density map estimation, localization, and tracking dataset to date, see Table \ref{tab:dataset-comparison}.

\begin{table*}[t]
\caption{Comparison between the DroneCrowd dataset and existing datasets.}
\centering
\setlength{\tabcolsep}{4pt}
\small
\begin{threeparttable}
\begin{tabular}{c|ccccccccc}
\hline
Dataset   &Type &Trajectory  &Resolution  &Frames &Max count &Min count &Ave count &Total count  &Year \\
\hline
UCF\_CC\_50 \cite{DBLP:conf/cvpr/IdreesSSS13}    &image  &  &-      &$50$      &$4,543$      &$94$          &$1,279.5$  &$63,974$  &2013 \\
Shanghaitech A \cite{DBLP:conf/cvpr/ZhangZCGM16} &image & &-     &$482$    &$3,139$ &$33$    &$501.4$       &$241,677$   &2016 \\
Shanghaitech B \cite{DBLP:conf/cvpr/ZhangZCGM16} &image & &$768\times1024$     &$716$ &$578$  &$9$   &$123.6$  &$88,488$ &2016 \\
AHU-Crowd \cite{DBLP:journals/jvcir/HuCNWL16}       &image  &  &$576\times720$   &$107$ &$2,201$   &$58$    &$420.6$      &$45,000$ &2016 \\
CARPK \cite{DBLP:conf/iccv/HsiehLH17}      &image  &  &$1280\times720$   &$1,448$ &$188$   &$1$            &$62.0$      &$89,777$ &2017 \\
Smart-City  \cite{DBLP:conf/wacv/ZhangSC18}      &image &   &$1920\times1080$   &$50$ &$14$   &$1$            &$7.4$      &$369$ &2018 \\
UCF-QNRF \cite{DBLP:conf/eccv/IdreesTAZARS18}    &image &  &- &$1,535$   &$12,865$     &$49$      &$815.4$      &$1,251,642$  &2018 \\
NWPU \cite{DBLP:journals/corr/abs-2001-03360} &image& &$2191\times3209$   &$5,109$ &$20,033$    &$0$    &$418.0$      &$2,133,375$ &2020 \\
\hline
\hline
UCSD \cite{DBLP:conf/cvpr/ChanLV08}               &video &  &$158\times238$   &$2,000$ &$46$          &$11$         &$24.9$       &$49,885$  &2008 \\
Mall \cite{DBLP:conf/iccv/LoyGX13}       &video  &  &$640\times480$   &$2,000$ &$53$   &$13$       &$31.2$      &$62,316$ &2013 \\
WorldExpo  \cite{DBLP:conf/cvpr/ZhangLWY15}      &video  &  &$576\times720$   &$3,980$ &$253$   &$1$            &$50.2$      &$199,923$ &2015 \\
FDST \cite{DBLP:conf/icmcs/FangZCGH19} &video &   &$1920\times1080$   &$15,000$ &$57$   &$9$            &$26.7$      &$394,081$ &2019 \\
\hline
\hline
DroneCrowd          &video &$\checkmark$ &$1920\times1080$ &$33,600$ &$455$ &$25$ &$144.8$ &$4,864,280$  &2021 \\
\hline
\end{tabular}
\end{threeparttable}
\label{tab:dataset-comparison}
\end{table*}

To handle this challenging dataset, we design a Space-Time Neighbor-Aware Network (STNNet) as a strong baseline, which solves the density map estimation, localization, and tracking simultaneously. Specifically, the proposed STNNet is formed by four modules, \ie, the feature extraction subnetwork, followed by the density map estimation heads, the localization, and the association subnets. The feature extraction subnetwork first uses two-branch CNNs to extract multi-scale features, and then computes the correlations between the extracted features in consecutive two frames to exploit the temporal relations. Using density map estimation heads, we can estimate the density of objects in video frames to perform crowd counting. Inspired by object detection \cite{DBLP:conf/nips/RenHGS15,DBLP:journals/pami/RenHG017,DBLP:conf/cvpr/ZhangWBLL18}, we introduce the localization subnet, formed by the classification and regression branches, to output accurate locations of targets in each individual frames. To exploit the temporal consistency, the association subnet is designed to predict motion offests of targets in consecutive frames for tracking. Besides, we develop the neighboring context loss by integrating spatial-temporal context of neighboring targets to guide the training of association subnet. Specifically, the neighboring context loss penalizes large displacements of the relative positions of adjacent objects in temporal domain, and guides the association subnet to generate accurate motion offsets. The whole network is trained in an end-to-end manner with the multi-task loss and Adam optimizer \cite{DBLP:journals/corr/KingmaB14}. After that, multi-object tracking methods \cite{DBLP:conf/cvpr/PirsiavashRF11,DBLP:conf/cvpr/AlahiGRRLS16} are used to predict long trajectories of targets. Compared with $12$ state-of-the-art algorithms, extensive experiments on our DroneCrowd dataset demonstrate the effectiveness of the proposed STNNet method for density map estimation, crowd localization and tracking tasks.

{\noindent {\bf Contributions.}} (1) We collect a large-scale drone captured dataset for density map estimation, localization, and tracking in dense crowd, which significantly surpasses existing datasets in terms of data type and volume, annotation quality, and difficulty. (2) We propose a space-time neighbor-aware network to solve the density map estimation, localization and tracking tasks simultaneously. (3) To exploit the spatial-temporal context, we design the neighboring context loss to penalize large displacements of the relative positions of adjacent objects in temporal domain for network training.

\begin{figure*}[t]
\centering
\includegraphics[width=1.0\linewidth]{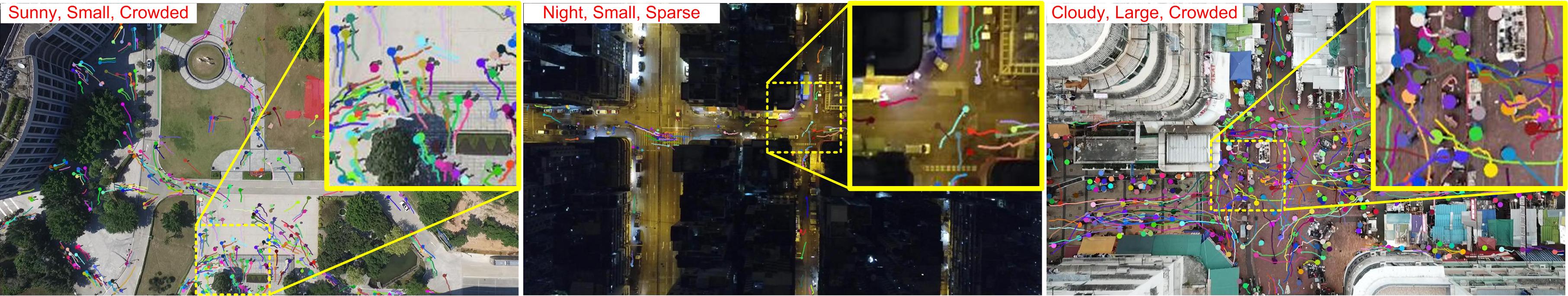}
\caption{Some annotated example frames in the DroneCrowd dataset. Different color indicates different object instance and the corresponding trajectory. The video-level attributes are presented on the top-left corner in each video frame.}
\label{fig:annotations}
\end{figure*}

\section{Related Work}
{\noindent {\bf Existing datasets.}} 
To date, there only exists a handful of crowd counting, crowd localization, or crowd tracking datasets. UCF\_CC\_50 \cite{DBLP:conf/cvpr/IdreesSSS13} is formed by $50$ images containing $64,000$ annotated humans, with the head counts ranging from $94$ to $4,543$. Shanghaitech \cite{DBLP:conf/cvpr/ZhangZCGM16} includes $1,198$ images with a total number of $330,165$ labeled people. Recently, UCF-QNRF \cite{DBLP:conf/eccv/IdreesTAZARS18} is released with $1,535$ images and $1.25$ million annotated people's heads in various scenarios. Hsieh~\etal~\cite{DBLP:conf/iccv/HsiehLH17} present a drone-based car counting dataset, which approximately contains $90K$ cars captured in different parking lots. Recently, Wang~\etal~\cite{DBLP:journals/corr/abs-2001-03360} collect a large-scale congested crowd counting and localization dataset, which includes more than $5K$ images and $2$ million annotated heads with points and boxes. However, these datasets are still limited in sizes and scenarios covered.

To evaluate counting algorithms in videos, Chan~\etal~\cite{DBLP:conf/cvpr/ChanLV08} present the UCSD counting dataset including low density crowd and counting difficulty. Similar to the UCSD dataset, Mall \cite{DBLP:conf/iccv/LoyGX13} is collected by the surveillance camera in a single location. Zhang~\etal~\cite{DBLP:conf/cvpr/ZhangLWY15} present the WorldExpo dataset with $3,980$ annotated frames in total, which is captured in $108$ different scenes during 2010 Shanghai WorldExpo. Fang~\etal~\cite{DBLP:conf/icmcs/FangZCGH19} collect a video dataset with $15K$ frames and $394K$ annotated heads captured from $13$ different scenes. In contrast to the aforementioned datasets, our DroneCrowd dataset is a large-scale drone-captured dataset for density map estimation, crowd localization and tracking, which consists of $112$ sequences with more than $4.8$ million head annotations on $20,800$ people trajectories.

{\noindent {\bf Crowd counting and density map estimation.}}
Modern crowd counting methods \cite{DBLP:conf/nips/LempitskyZ10,DBLP:conf/cvpr/ZhangZCGM16,DBLP:conf/cvpr/SamSB17,DBLP:conf/eccv/CaoWZS18,DBLP:conf/cvpr/LiZC18,DBLP:conf/cvpr/LiuSF19,DBLP:conf/aaai/LuoYLNJZC20,DBLP:conf/nips/WangLSH20} formulate crowding counting as density map estimation. Lempitsky and Zisserman \cite{DBLP:conf/nips/LempitskyZ10} learn to infer the density estimation by a minimization of a regularized risk quadratic cost function. Zhang~\etal~\cite{DBLP:conf/cvpr/ZhangZCGM16} use the multi-column CNN network to estimate the crowd density map, which learns the features for different head sizes by each column CNN. Sam~\etal~\cite{DBLP:conf/cvpr/SamSB17} develop the switching CNN model to handle the variations of crowd density. Cao~\etal~\cite{DBLP:conf/eccv/CaoWZS18} propose an encoder-decoder network, where the encoder extracts multi-scale features with scale aggregation and the decoder generates high-resolution density maps using transposed convolutions. Li~\etal~\cite{DBLP:conf/cvpr/LiZC18} employ dilated convolution layers to enlarge receptive fields and extract deeper features without losing resolutions. Liu~\etal~\cite{DBLP:conf/cvpr/LiuSF19} adaptively encodes the scale of the contextual information for accurate crowd density prediction. In \cite{DBLP:conf/iros/LiuLSF19}, the physically-inspired temporal consistency constraints are considered in the network to handle the viewpoint changes by drones. Besides, Luo~\etal~\cite{DBLP:conf/aaai/LuoYLNJZC20} propose the hybrid graph neural network to capture dependencies among multi-scale counting and localization features. To avoid hurting the generalization bound of a model, Wang~\etal~\cite{DBLP:conf/nips/WangLSH20} propose the optimal transport to measure the similarity between the normalized predicted and ground-truth density maps.

In terms of crowd counting in videos, spatio-temporal information is critical to improve the counting accuracy. Xiong~\etal~\cite{DBLP:conf/iccv/XiongSY17} design a convolutional LSTM model to fully capture both spatial and temporal dependencies for crowd counting. Zhang~\etal~\cite{DBLP:conf/iccv/ZhangWCM17} combine fully convolutional neural networks and LSTM by residual learning to perform vehicle counting. Liu~\etal~\cite{DBLP:conf/eccv/LiuSF20} first compute people flows between consecutive frames and then estimate the densities from these flows. Different from existing methods, our STNNet can output both crowd density and target locations in crowds using the proposed localization subnet.

{\noindent \textbf{Crowd localization and tracking.}} 
Besides crowd counting, crowd localization and tracking are also important tasks in safety control scenarios. Rodriguez~\etal~\cite{DBLP:conf/iccv/RodriguezLSA11} formulate an energy minimization framework by jointly optimizing the density and location, with the temporal-spatial constraints of person tracks in video. Ma~\etal~\cite{DBLP:conf/cvpr/MaYC15} first obtain local counts from sliding windows over the density map and then use integer programming to recover the locations of individual objects. In \cite{DBLP:conf/eccv/IdreesTAZARS18}, crowd counting and localization tasks are simultaneously solved with a CNN model trained by a composition loss. In contrast, our method captures context information among neighbouring targets and estimate motion offsets of targets between consecutive frames, trained by the proposed neighboring context loss.

\begin{figure*}[t]
\centering
\includegraphics[width=.95\linewidth]{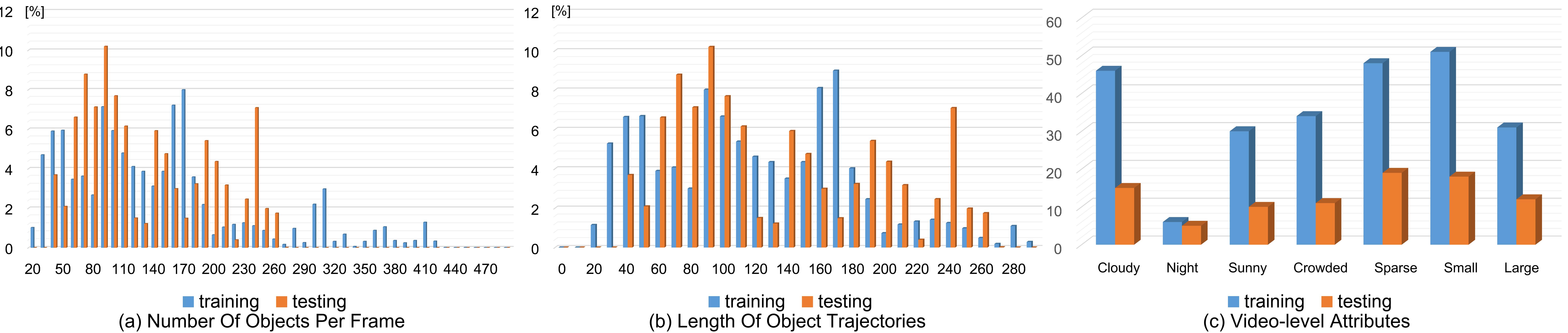}
\caption{(a) The distribution of the number of objects per frame, (b) the distribution of the length of object trajectories, and (c) the attribute statistics, of the {\tt training} and {\tt testing} sets in the DroneCrowd dataset.}
\label{fig:attributes}
\end{figure*}

\begin{table}[t]
  \centering
  \setlength{\tabcolsep}{3pt}
  \caption{Statistics of each attribute in DroneCrowd.}
  \small{
    \begin{tabular}{c|cccc}
    \hline    
    Attribute &{Min count}  &{Max count} &{Avg count} &{Frames} \\
    \hline
    Small  & $26$    & $455$   & $143.8$ & $20,700$ \\
    Large   & $25$    & $436$   & $146.3$ & $12,900$ \\
    \hline
    Cloudy & $25$    & $436$   & $144.9$ & $18,300$ \\
    Sunny & $26$    & $455$   & $153.3$ & $12,300$ \\
    Night & $40$    & $167$   & $109.1$ & $3,000$ \\
    \hline
    Crowded & $129$   & $455$   & $225.6$ & $13,500$ \\
    Sparse & $25$    & $170$   & $90.5$  & $20,100$ \\
    \hline
    \end{tabular}}
  \label{tab:attribute-stats}
\end{table}

\section{DroneCrowd Dataset}
\subsection{Data Collection and Annotation}\label{sec:dataset}
Our DroneCrowd dataset is captured by drone-mounted cameras (\ie, DJI Phantom 4, Phantom 4 Pro and Mavic), covering a wide range of scenarios, \eg, campus, street, park, parking lot, playground and plaza\footnote{We strictly comply with local laws and regulations in China when using unmanned aircraft/drones, and avoid restricted areas to capture videos. Since the scales of objects are extremely small, no identity information such as faces and vehicle plates could be retrieved. After careful check, we confirm that all data in our dataset would not leak any personal information.}. The videos are recorded at $25$ frames per seconds (FPS) with a resolution of $1920\times1080$ pixels. As presented in Figure \ref{fig:attributes} (a) and (b), the maximal and minimal numbers of people in each video frame are $455$ and $25$ respectively, and the average number of objects is $144.8$. Moreover, more than $20$ thousands of head trajectories of people are annotated with more than $4.8$ million head points in individual frames of $112$ video clips. Over $20$ domain experts annotate and double-check the dataset using the vatic software \cite{DBLP:journals/ijcv/VondrickPR13} for more than two months. Figure \ref{fig:annotations} shows some frames of video clips with annotated trajectories of people heads.

We divide DroneCrowd into the {\tt training} and {\tt testing} sets, with $82$ and $30$ sequences, respectively. Notably, training videos are taken at different locations from testing videos to reduce the chances of algorithms to overfit to particular scenes. It contains video sequences with large variations in scale, viewpoint, and background clutters. To analyze the performance of algorithms thoroughly, we define three video-level attributes of the dataset, described as follows. (1) \textit{Illumination}: under different illumination conditions, the objects are assumed to be different in appearance. Three categories of illumination conditions are considered in our dataset, including {\it Cloudy}, {\it Sunny}, and {\it Night}. (2) \textit{Scale} indicates the size of objects. Two categories of scales are defined, including {\it Large} (the diameter of objects $>15$ pixels) and {\it Small} (the diameter of objects $\leq15$ pixels). (3) \textit{Density} indicates the number of objects in each frame. Based on the average number of objects in each frame, we divide the dataset into two density levels, \ie, {\it Crowded} (with the number of objects in each frame larger than $150$), and {\it Sparse} (with the number of objects in each frame less than $150$). The statistics on different attributes are shown in Figure \ref{fig:attributes} (c) and Table \ref{tab:attribute-stats}.

\subsection{Evaluation Metrics and Protocols}
{\noindent \textbf{Density map estimation.}}
Following the previous works \cite{DBLP:conf/cvpr/ZhangLWY15,DBLP:conf/cvpr/ZhangZCGM16,DBLP:conf/eccv/IdreesTAZARS18}, the density map estimation task aims to compute per-pixel density at each location in the image, while preserving spatial information about distribution of people. We use the mean absolute error (\text{MAE}) and mean squared error (\text{MSE}) to evaluate the performance, \ie, $\text{MAE} = \frac{1}{\sum_{i=1}^{M}N_i}\sum_{i=1}^{M}\sum_{j=1}^{N_i}{|z_{i,j}-\hat{z}_{i,j}|}$, and  $\text{MSE} = \sqrt{\frac{1}{\sum_{i=1}^{M}N_i}\sum_{i=1}^{M}\sum_{j=1}^{N_i}{|z_{i,j}-\hat{z}_{i,j}|^2}}$, where $M$ is the number of video clips, $N_i$ is the number of frames in the $i$-th video. $z_{i,j}$ and $\hat{z}_{i,j}$ are the ground-truth and estimated number of people in the $j$-th frame of the $i$-th video clip, respectively. As stated in \cite{DBLP:conf/cvpr/ZhangZCGM16}, \text{MAE} and \text{MSE} describe the accuracy and robustness of the estimation respectively.

{\noindent \textbf{Crowd localization.}}
The goal of crowd localization is to detect the locations of all people in an image. Each evaluated crowd localization algorithm is required to output a series of detected points with confidence scores for each test image. The estimated locations determined by the confidence threshold are associated to the ground-truth locations using greedy method. Then, we compute the L-mAP at various distance thresholds ($1,2,3,\cdots,25$ pixels) to evaluate the localization results. We also report the performance with three specific distance thresholds, \ie, L-AP$@10$, L-AP$@15$, and L-AP$@20$ pixels. These criteria penalize missing detection of people as well as duplicate detections.

{\noindent \textbf{Crowd tracking.}}
Crowd tracking requires an algorithm to recover the trajectories of people in video sequence, which is evaluated on the metric in \cite{isvrc-2017}. Specifically, each tracker is required to output a series of head points with confidence scores and the corresponding identities. We sort the tracklets, formed by the locations with the same identity, based on the average confidence of their detections. A tracklet is considered to be correct if the matched ratio between the predictions and ground-truth tracklets is larger than a threshold. We use $3$ thresholds in evaluation, \ie, $0.10$, $0.15$, and $0.20$. The matching distance threshold between the predicted and ground-truth locations on the tracklets is set to $25$ pixels. The T-mAP scores over different thresholds (\ie, T-AP$@0.10$, T-AP$@0.15$, and T-AP$@0.20$) are used to measure the performance. 
\begin{figure*}[t]
\centering
\includegraphics[width=0.9\linewidth]{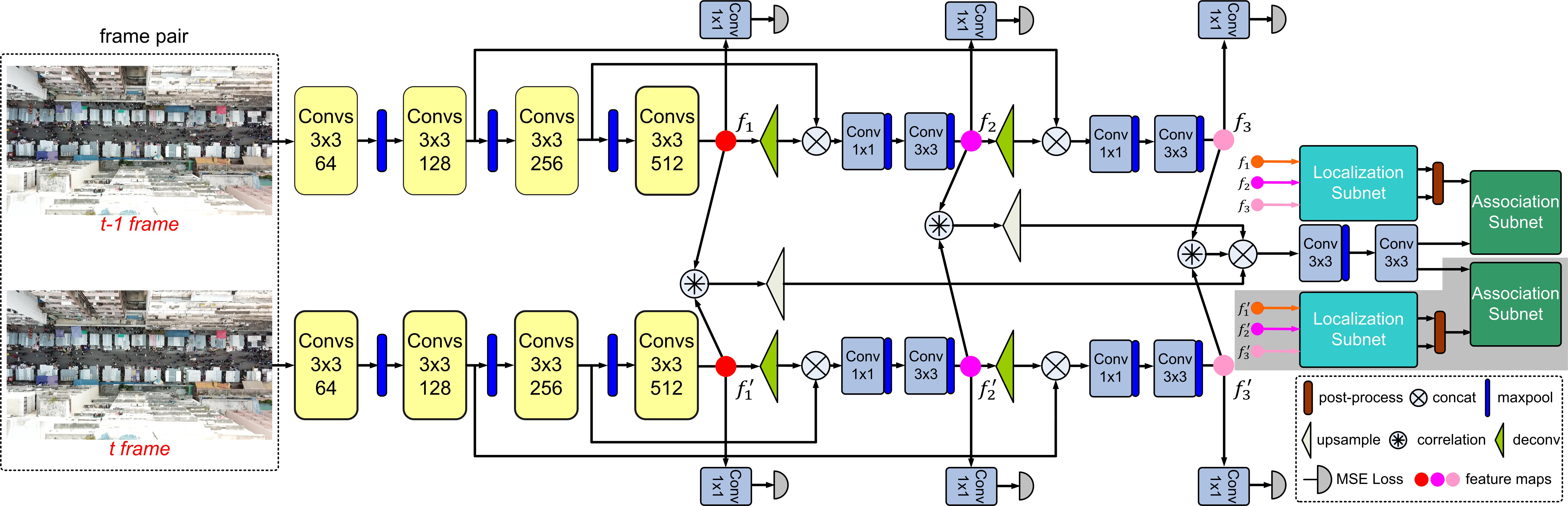}
\caption{The architecture of our STNNet. The yellow rectangles indicate the convolution groups in the VGG-16 backbone. The blue and green rectangles indicate the localization subnet (see Figure \ref{fig:localization}) and association subnet (see Figure \ref{fig:association}) respectively. Colourful circles indicate feature maps at different stage. Note that the modules in the grey regions are removed in the testing phase.}
\label{fig:architecture}
\end{figure*}

\section{Our Method}
Our STNNet sequentially takes a pair of frames as input, and outputs the density maps, the locations, and the motion offsets of objects in these two frames, see Figure \ref{fig:architecture}. After that, the association method is used to generate long trajectories of objects in videos.

\subsection{Network Architecture} 
As shown in Figure \ref{fig:architecture}, the Siamese feature extraction subnetwork in our STNNet is constructed on the first $4$ groups of convolution layers in the parameters shared two-branch VGG-16 network \cite{DBLP:journals/corr/SimonyanZ14a} to extract multi-scale features. Inspired by \cite{DBLP:conf/miccai/RonnebergerFB15}, the U-Net style architecture is used to fuse multi-scale features for prediction. Using density map estimation heads, we can determine the number of targets based on multi-scale features. Meanwhile, the correlation operation \cite{DBLP:conf/cvpr/IlgMSKDB17} is conducted on the extracted features to exploit the temporal coherence at different stage. In addition, the localization and association subnets are introduced to predict the locations of target points and the corresponding motion offsets, which are described as follows. 

\begin{figure}[t]
\centering
\includegraphics[width=.9\linewidth]{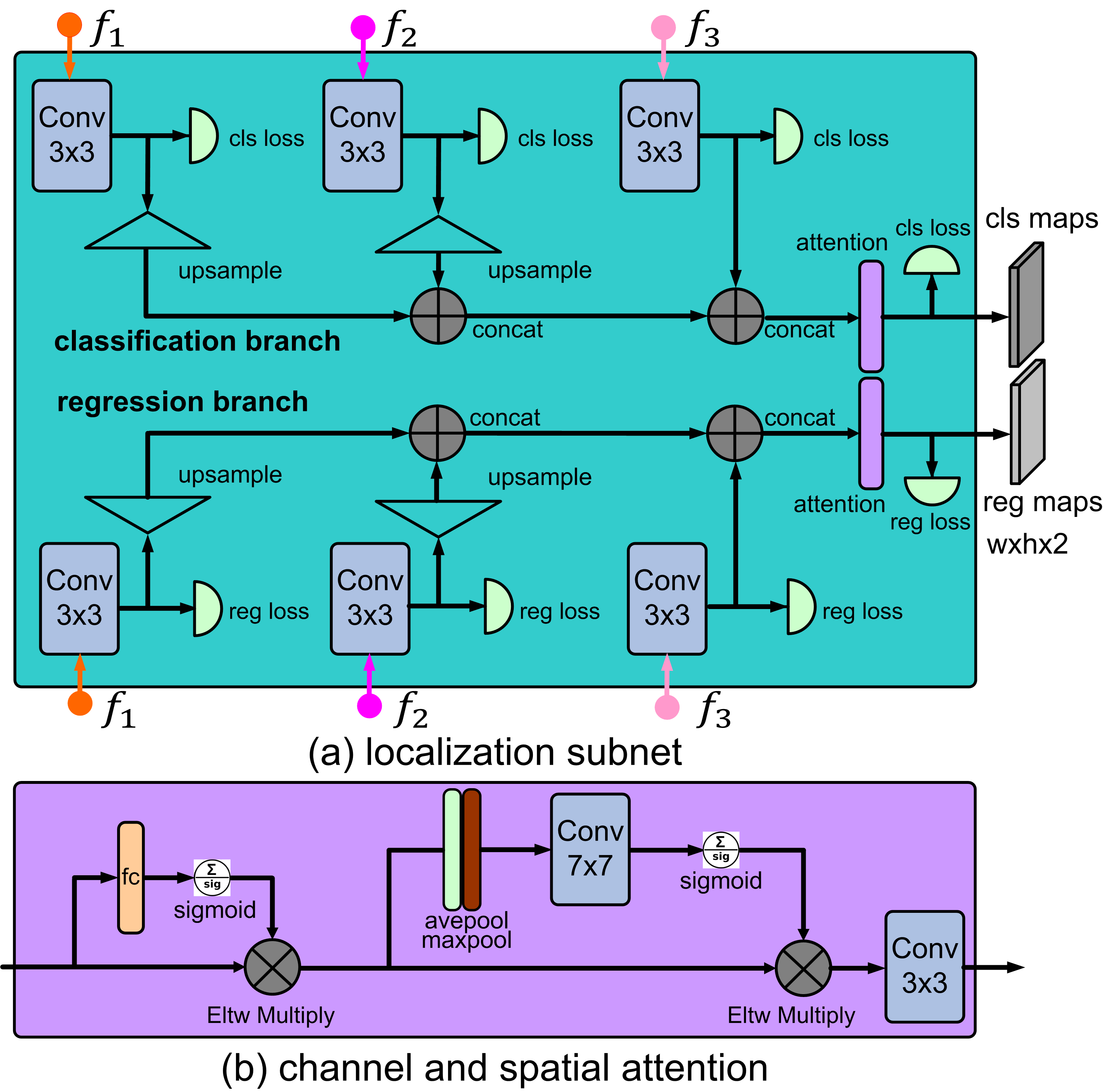}
\caption{(a) the localization subnet based on (b) channel and spatial attention.}
\label{fig:localization}
\end{figure}

{\noindent \textbf{Localization subnet.}}
The localization subnet consists of the classification and regression branches. To generate accurate locations of objects, we tile the object proposal in each pixel. The classification branch aims to predict the probability of each proposal to be an object, and the regression branch aims to generate the accurate locations of the positive proposals. As shown in Figure \ref{fig:localization}, we fuse multi-scale feature maps (\ie, $f_1$, $f_2$, $f_3$) with both channel and spatial attention \cite{DBLP:conf/eccv/WooPLK18} for each branch. After that, we resize multi-scale feature maps and then output the fused classification and regression maps. The classification map denotes the probability that each proposal contains an object and the regression map contains the regressed offsets of the positive proposals. Finally, we perform non-maximal suppression to predict the accurate object locations.
\begin{figure*}[t]
\centering
\includegraphics[width=.9\linewidth]{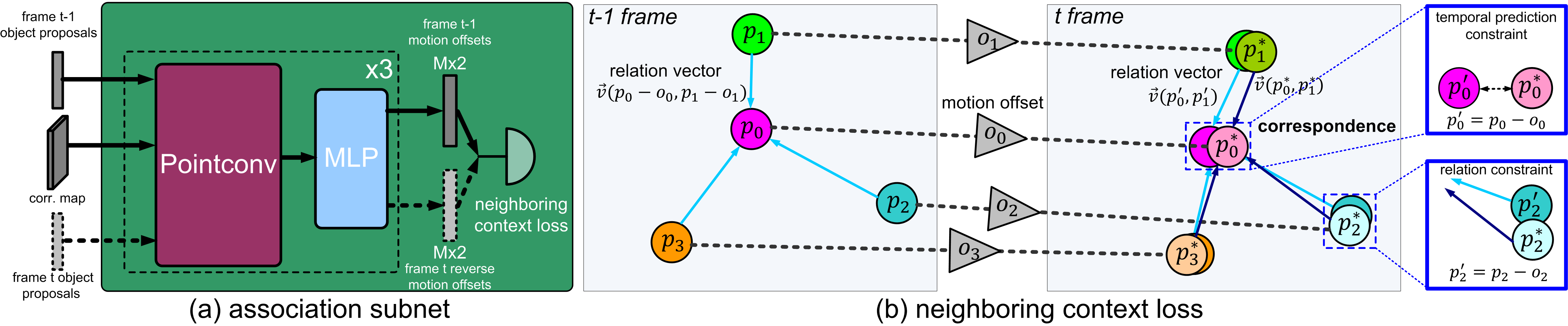}
\caption{(a) the association subnet using (b) the neighboring context loss. Notably, the dashed modules in (a) are only used in the training phase. For clarity, we only display the calculation of the terms from time $t-1$ to time $t$ in the neighboring context loss.}
\label{fig:association}
\end{figure*}

{\noindent \textbf{Association subnet.}}
As mentioned above, we introduce the association subnet to predict the motion offsets of each object to complete the tracking task. As shown in Figure \ref{fig:association}(a), given the $M$ top scored post-processed object proposals generated by the localization subnet in the $(t-1)$-th frame and the fused multi-scale correlation features, we use $3$ stacked PointConv \cite{DBLP:conf/cvpr/WuQL19} and multi-layer perceptron (MLP) operations to construct the association subnet to generate the motion offsets in a circle, \ie, from the $(t-1)$-th frame to $t$-th frame and vice versa. Note that, only the nearest $\beta$ points are considered in each PointConv operation. 

\subsection{Multi-Task Loss Function}
We use the multi-task loss to guide the training of our STNNet method, which consists of three terms, including the neighboring context loss ${\it L}_\text{ass}(\hat{\it P}_k, {\it P}^\ast_k, {\it O}_k)$, the localization loss ${\it L}_\text{loc}(\hat{{\it C}}_k, {\it C}^\ast_k, \hat{{\it R}}_k, {\it R}^\ast_k)$, and the density loss ${\it L}_\text{den}(\hat{\Phi}_{k}, \Phi^\ast_{k})$, \ie, 
\begin{equation}
\begin{array}{ll}
{\cal L} &= \frac{1}{{\it K}}\sum_{k=1}^{{\it K}}\big({\it L}_\text{den}(\hat{\Phi}_{k}, \Phi^\ast_{k})
+{\it L}_\text{loc}(\hat{{\it C}}_k, {\it C}^\ast_k, \hat{{\it R}}_k, {\it R}^\ast_k)\\ 
&+\sum_{k}^{\it K}{\it L}_\text{ass}(\hat{\it P}_k, {\it P}^\ast_k, {\it O}_k)\big),
\end{array}
\label{equ:multi-task-loss}
\end{equation}
where $k$ is the index of the batch, ${\it K}$ is the batch size. $\hat{\Phi}_{k}$ and $\Phi^\ast_{k}$ are the predicted and ground-truth density maps. $\hat{{\it C}}_k$ and ${\it C}^\ast_k$ are the predicted and ground-truth labels (\ie, objects or background) of the object proposals, $\hat{{\it R}}_k$ and ${\it R}^\ast_k$ are the predicted and ground-truth offsets of the object proposals. $\hat{\it P}_k$ and ${\it P}^\ast_k$ are the predicted and ground-truth locations of the objects, and ${\it O}_k$ is the prediction motion offsets of the objects. In the following sections, we would like to discuss each loss term in details. 

{\noindent \textbf{Density loss.}} 
Inspired by \cite{DBLP:conf/cvpr/ZhangZCGM16}, we use the pixel-wise Euclidean loss for the density loss. The geometry-adaptive Gaussian kernel method is used to generate the ground-truth density map $\Phi^\ast_{k}$. The density loss term is computed as
\begin{equation}
\begin{array}{ll}
&{\it L}_{\text{den}}(\hat{\Phi}_{k}, \Phi^\ast_{k})= \frac{1}{2\cdot {L}}\sum_{t=1}^{2}\sum_{l=1}^{L}\sum_{i=1}^{W_{l}} \\
& \sum_{j=1}^{H_{l}}\omega_{l}\cdot\| \hat{\phi}_{k,t}(i,j,l) - \phi^\ast_{k,t}(i,j,l) \|^2_2,
\end{array}
\label{equ:loss-counting}
\end{equation}
where $\hat{\phi}_{k,t}(i,j,l)$ and $\phi^\ast_{k,t}(i,j,l)$ are the values of the predicted and ground-truth density maps at $(i, j)$ of layer $l$ at time $t$ of the $k$-th batch, and $\omega_{l}$ is the parameter to balance the influence of each layer. 

{\noindent \textbf{Localization loss.}} 
Motivated by object detection \cite{DBLP:conf/nips/RenHGS15,DBLP:journals/pami/RenHG017,DBLP:conf/cvpr/ZhangWBLL18}, the localization loss is formed by the classification loss and regression loss. We tile the point proposals on each pixel and match the proposal to the ground-truth points. If the proposal locates in the neighboring regions of the ground-truth points, we assign it to be the positive proposal (\ie, $s_{k}(i,j,l)=1$ for the proposal at $(i, j)$ in the $l$-th layer in the $k$-th batch); otherwise the background (\ie, $s_{k}(i,j,l)=0$). Thus, the localization loss is computed as 
\begin{equation}
\begin{array}{ll}
& {\it L}_{\text{loc}}(\hat{{\it C}}_k, {\it C}^\ast_k, \hat{{\it R}}_k, {\it R}^\ast_k) \\
&=\frac{1}{L}\sum_{l=1}^{L}\sum_{i=1}^{W_{l}}\sum_{j=1}^{H_{l}}\big(L_\text{cls}(\hat{\it c}_k(i,j,l),{\it c}^\ast_k(i,j,l)), \\
&+ s_{k}(i,j,l) \cdot L_\text{reg}(\hat{\it r}_k(i,j,l),{\it r}^\ast_k(i,j,l))\big),
\end{array}
\label{equ:loss-localization}
\end{equation}
where $\hat{\it c}_k(i,j,l)$ and ${\it c}^\ast_k(i,j,l)$ are the predicted and ground-truth labels at $(i, j)$ of layer $l$. $\hat{\it r}_k(i, j, l)$ and ${\it r}^\ast_k(i, j, l)$ are the predicted and ground-truth offsets at $(i, j)$ of layer $l$. We use the log loss to compute $L_\text{cls}$, and the squared loss to compute $L_\text{reg}$. Notably, the regression loss $L_\text{reg}$ is only activated for the positive proposals.  

{\noindent \textbf{Neighboring context loss.}} 
In crowded scenes, the objects are generally clustered in a small region and usually share similar motion patterns in consecutive frames. To exploit the motion consistency of neighboring objects, we design a neighboring context loss, which is formed by two parts, \ie, the temporal prediction constraint, and the relation constraint, see Figure \ref{fig:association}(b).

Specifically, the temporal prediction constraint enforces the proposals in the consecutive frames projected by the predicted motion offsets to approach the ground-truth points. Let $p_{i,t-1}$ be the location of the $i$-th proposal at time $t-1$, $p_{j, t-1}\in{\cal N}_{p_{i,t-1}}$ be the object location in the neighboring region of the proposal at $p_{i,t-1}$, and $o_{i,t-1}$ be the predicted offset corresponding to the proposal at $p_{i,t-1}$ from time $t-1$ to $t$. Thus, the temporal prediction constraint aims to minimize the $\ell_1$-norm of the differences, \ie, $\|(p_{i,t-1}-o_{i,t-1})-p^\ast_{i,t}\|_{1}$. Meanwhile, the relation constraint enforces the relation vectors between the target and neighboring objects to approach to the relation vectors of their corresponding associated ground-truth points. Let $\vec{v}(p_{i,t-1}-o_{i,t-1}, p_{j,t-1}-o_{j,t-1})$ be the relation vector\footnote{The relation vector is computed as $\vec{v}(p_{i,t-1}-o_{i,t-1}, p_{j,t-1}-o_{j,t-1})=(p_{j,t-1}-o_{j,t-1})-(p_{i,t-1}-o_{i,t-1})$.} between the target and neighboring objects projected to the second frame, and $\vec{v}(p^\ast_{i,t}, p^\ast_{j,t})$ be the relation vector between the ground-truth points at $p^\ast_{i,t}$ and $p^\ast_{j,t}$. Thus, the relation constraint aims to minimize $\sum_{p_{j,t-1}\in{\mathcal{N}_{p_{i,t-1}}}}\|\vec{v}(p_{i,t-1}-o_{i,t-1}, p_{j,t-1}-o_{j,t-1}) - \vec{v}(p^\ast_{i,t}, p^\ast_{j,t})\|_{1}$. The cycle strategy is used to compute the neighboring context loss, \ie, 
\begin{equation}
\begin{array}{ll}
&{\it L}_\text{ass}(\hat{\it P}_k, {\it P}^\ast_k, {\it O}_k)= \frac{1}{2\cdot M}\sum_{i=1}^{M}\big(\|p'_{i,t-1}-p^\ast_{i,t}\|_{1}\\
&+\sum_{p_{j,t-1}\in{\mathcal{N}_{p_{i,t-1}}}}\|\vec{v}(p'_{i,t-1}, p'_{j,t-1})- \vec{v}(p^\ast_{i,t}, p^\ast_{j,t})\|_{1}\\
&+\|p'_{i,t}-p^\ast_{i,t-1}\|_{1}\\
&+\sum_{p_{j,t}\in{\mathcal{N}_{p_{i,t}}}}\|\vec{v}(p'_{i,t}, p'_{j,t})- \vec{v}(p^\ast_{i,t-1}, p^\ast_{j,t-1})\|_{1}
\big),
\end{array}
\label{equ:loss-association}
\end{equation}
where $p'_{i,t}=p_{i,t}-o_{i,t}$ and $p'_{j,t}=p_{j,t}-o_{j,t}$ are the projected targets.

\begin{figure*}[t]
\centering
\includegraphics[width=0.95\linewidth]{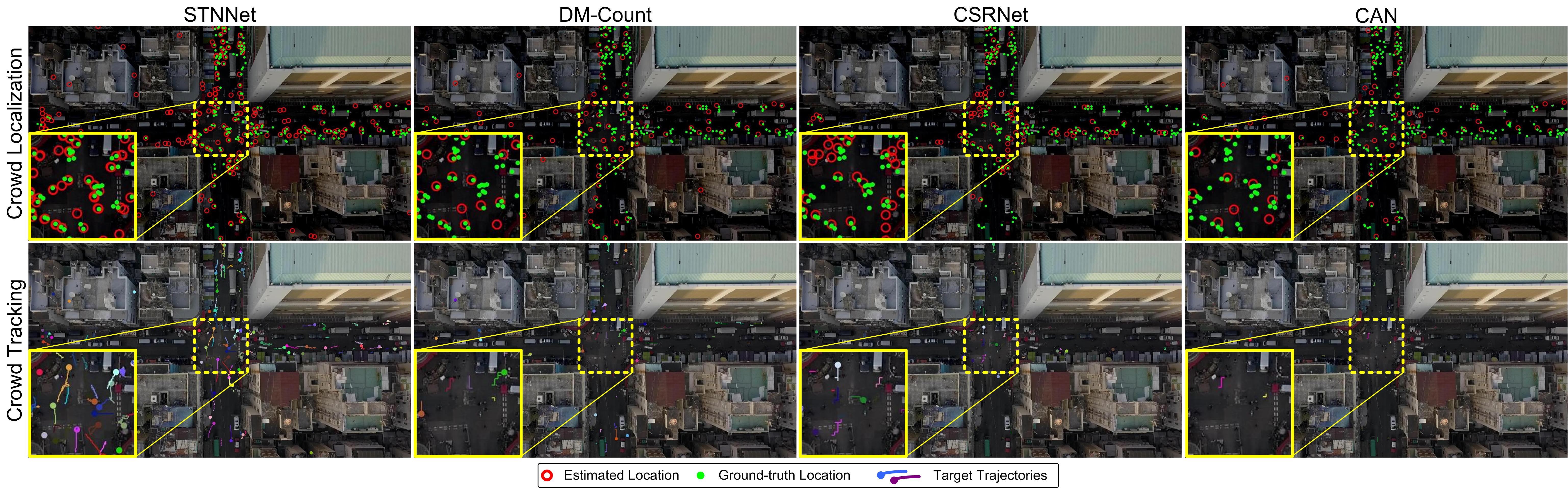}
\caption{Qualitative results of DM-Count \cite{DBLP:conf/nips/WangLSH20}, CSRNet \cite{DBLP:conf/cvpr/LiZC18}, CAN \cite{DBLP:conf/cvpr/LiuSF19}, and our STNNet on DroneCrowd. Best view in color version.}
\label{fig:crowd-visual-results}
\end{figure*}

\subsection{Optimization}
To increase diversity in training data, we randomly flip and crop the training images. Due to limited computation resources, we equally divide each frame into $2\times2$ patches, and use the divided $4$ patches with the resolution of $960\times540$ for training. For the Pointconv layer, we use $\beta=8$ nearest points to capture the context information. In \eqref{equ:loss-counting}, the pre-set weights $\omega_{l}$ are set to $\{2.0,0.5,0.05\}$. The matching threshold between the proposals and ground-truth points is set to $10$ pixels. Meanwhile, the threshold used to determine the neighboring regions of pixels $\mathcal{N}_{p_i}$ is set to $50$ pixels. The total number of proposal objects $M$ is set to $128$. In addition, we set the batch size $K=4$ in the training phase. 

{\noindent \textbf{Two-stage training.}} 
We use the two stage strategy to train our network. For the first stage, we remove the association subnet and train the network to generate accurate density map and object proposals. After that, we fixed the parameters in the density map estimation heads, and add the association subnet to fine-tune the whole network. We use the Adam optimization algorithm \cite{DBLP:journals/corr/KingmaB14} with the learning rate $10^{-6}$ in both stages. 

\section{Experiment}
As discussed above, we conduct the experiment on our DroneCrowd for crowd counting, localization and tracking. We report the density map estimation results and speeds of STNNet and $12$ existing methods. Meanwhile, the ablation study is conducted to verify the effectiveness of important components in our method. Besides, some visual results are shown in Figure \ref{fig:crowd-visual-results}. 

\begin{table*}[t]
\caption{Estimation errors of density maps on DroneCrowd.}
\centering
\setlength{\tabcolsep}{1.2pt}
\small{
\begin{tabular}{c|c||cc||cc|cc||cc|cc|cc||cc|cc}
\hline
\multirow{2}{*}{Method} &Speed &\multicolumn{2}{c||}{Overall} &\multicolumn{2}{c|}{Large} &\multicolumn{2}{c||}{Small} &\multicolumn{2}{c|}{Cloudy} &\multicolumn{2}{c|}{Sunny} &\multicolumn{2}{c||}{Night} &\multicolumn{2}{c|}{Crowded} &\multicolumn{2}{c}{Sparse}\\
\cline{3-18}
&FPS &$\text{MAE}$ &$\text{MSE}$ &$\text{MAE}$ &$\text{MSE}$ &$\text{MAE}$ &$\text{MSE}$ &$\text{MAE}$ &$\text{MSE}$ &$\text{MAE}$ &$\text{MSE}$ &$\text{MAE}$ &$\text{MSE}$ &$\text{MAE}$ &$\text{MSE}$ &$\text{MAE}$ &$\text{MSE}$\\
\hline
MCNN \cite{DBLP:conf/cvpr/ZhangZCGM16} &${\bf 28.98}$ &$34.7$ &$42.5$ &$36.8$ &$44.1$ &$31.7$ &$40.1$ &$21.0$ &$27.5$ &$39.0$ &$43.9$ &$67.2$ &$68.7$ &$29.5$ &$35.3$ &$37.7$ &$46.2$\\
C-MTL \cite{DBLP:conf/avss/SindagiP17} &$2.31$ &$56.7$ &$65.9$ &$53.5$ &$63.2$ &$61.5$ &$69.7$ &$59.5$ &$66.9$ &$56.6$ &$67.8$ &$48.2$ &$58.3$ &$81.6$ &$88.7$ &$42.2$ &$47.9$\\
MSCNN \cite{DBLP:conf/icip/ZengXCQZ17} &$1.76$ &$58.0$ &$75.2$ &$58.4$ &$77.9$ &$57.5$ &$71.1$ &$64.5$ &$85.8$ &$53.8$ &$65.5$ &$46.8$ &$57.3$ &$91.4$ &$106.4$ &$38.7$ &$48.8$\\
LCFCN \cite{DBLP:conf/eccv/LaradjiRPVS18} &$3.08$ &$136.9$ &$150.6$ &$126.3$ &$140.3$ &$152.8$ &$164.8$ &$147.1$ &$160.3$ &$137.1$ &$151.7$ &$105.6$ &$113.8$ &$208.5$ &$211.1$ &$95.4$ &$110.0$\\
SwitchCNN \cite{DBLP:conf/cvpr/SamSB17} &$0.01$ &$66.5$ &$77.8$ &$61.5$ &$74.2$ &$74.0$ &$83.0$ &$56.0$ &$63.4$ &$69.0$ &$80.9$ &$92.8$ &$105.8$ &$67.7$ &$79.8$ &$65.7$ &$76.7$\\
ACSCP \cite{DBLP:conf/cvpr/ShenXNWHY18} &$1.58$ &$48.1$ &$60.2$ &$57.0$ &$70.6$ &$34.8$ &$39.7$ &$42.5$ &$46.4$ &$37.3$ &$44.3$ &$86.6$ &$106.6$ &$36.0$ &$41.9$ &$55.1$ &$68.5$\\
AMDCN \cite{DBLP:conf/cvpr/DebV18} &$0.16$ &$165.6$ &$167.7$ &$166.7$ &$168.9$ &$163.8$ &$165.9$ &$160.5$ &$162.3$ &$174.8$ &$177.1$ &$162.3$ &$164.3$ &$165.5$ &$167.7$ &$165.6$ &$167.8$\\
StackPooling \cite{DBLP:journals/corr/abs-1808-07456} &$0.73$ &$68.8$ &$77.2$ &$68.7$ &$77.1$ &$68.8$ &$77.3$ &$66.5$ &$75.9$ &$74.0$ &$83.4$ &$65.2$ &$67.4$ &$95.7$ &$101.1$ &$53.1$ &$59.1$\\
DA-Net \cite{DBLP:journals/access/ZouSQZ18} &$2.52$ &$36.5$ &$47.3$ &$41.5$ &$54.7$ &$28.9$ &$33.1$ &$45.4$ &$58.6$ &$26.5$ &$31.3$ &$29.5$ &$34.0$ &$56.5$ &$68.3$ &$24.9$ &$28.7$\\
CSRNet \cite{DBLP:conf/cvpr/LiZC18} &$3.92$ &$19.8$ &$25.6$ &$17.8$ &$25.4$ &$22.9$ &$25.8$ &$12.8$ &$16.6$ &$19.1$ &$22.5$ &$42.3$ &$45.8$ &$20.2$ &$24.0$ &$19.6$ &$26.5$\\
CAN \cite{DBLP:conf/cvpr/LiuSF19} &$7.12$ &$22.1$ &$33.4$ &$18.9$ &$26.7$ &$26.9$ &$41.5$ &$\bf 11.2$ &$\bf 14.9$ &$14.8$ &$17.5$ &$69.4$ &$73.6$ &$\bf 14.4$ &$\bf 17.9$ &$26.6$ &$39.7$\\
DM-Count \cite{DBLP:conf/nips/WangLSH20} &$10.04$ &$18.4$ &$27.0$ &$19.2$ &$29.6$ &$17.2$ &$22.4$ &$11.4$ &$16.3$ &$\bf 12.6$ &$\bf 15.2$ &$51.1$ &$55.7$ &$17.6$ &$21.8$ &$18.9$ &$29.6$\\
\hline
\hline
STNNet (w/o loc) &$3.65$ &$18.6$ &$22.2$ &$17.1$ &$20.5$ &$21.0$ &$24.6$ &$14.7$ &$19.9$ &$21.4$ &$23.3$  &$24.7$ &$26.3$ &$24.2$ &$27.3$ &$15.4$ &$18.7$\\
STNNet &$3.41$ &$\bf 15.8$ &$\bf 18.7$ &$\bf 16.0$ &$\bf 18.4$ &$\bf 15.6$ &$\bf 19.2$ &$14.1$ &$17.2$ &$19.9$ &$22.5$  &$\bf 12.9$ &$\bf 14.4$ &$18.5$ &$21.6$ &$\bf 14.3$ &$\bf 16.9$\\
\hline
\end{tabular}}
\label{tab:density-map-estimation-results}
\end{table*}

\begin{table*}[t]
\begin{minipage}{0.48\linewidth}
\centering
\caption{Localization accuracy on DroneCrowd.}
\setlength{\tabcolsep}{1.0pt}
\small{
\begin{tabular}{c|c|ccc}
\hline
Methods &L-mAP &L-AP$@10$ &L-AP$@15$ &L-AP$@20$ \\
\hline
MCNN \cite{DBLP:conf/cvpr/ZhangZCGM16} &$9.05\%$ &$9.81\%$ &$11.81\%$ &$12.83\%$ \\
CAN \cite{DBLP:conf/cvpr/LiuSF19} &$11.12\%$ &$8.94\%$ &$15.22\%$ &$18.27\%$ \\
CSRNet \cite{DBLP:conf/cvpr/LiZC18} &$14.40\%$ &$15.13\%$ &$19.77\%$ &$21.16\%$ \\
DM-Count \cite{DBLP:conf/nips/WangLSH20} &$18.17\%$ &$17.90\%$ &$25.32\%$ &$27.59\%$ \\
\hline\hline
STNNet (w/o loc) &$32.19\%$  &$33.88\%$ &$39.56\%$ &$43.22\%$ \\
STNNet (w/o ass) &$39.77\%$  &$42.06\%$ &$50.00\%$ &$54.88\%$ \\
STNNet (w/o rel) &$40.00\%$  &$42.29\%$ &$50.31\%$ &$55.11\%$ \\
STNNet (w/o cyc) &$40.23\%$  &$42.57\%$ &$50.64\%$ &$55.42\%$ \\
STNNet &${\bf 40.45\%}$  &${\bf 42.75\%}$ &${\bf 50.98\%}$ &${\bf 55.77\%}$ \\
\hline
\end{tabular}}
\label{tab:crowd-localization-results}
\end{minipage}
\begin{minipage}{0.52\linewidth}
\centering
\caption{Tracking accuracy on DroneCrowd in terms of min-cost flow/social-LSTM.}
\setlength{\tabcolsep}{1pt}
\footnotesize{
\begin{tabular}{c|c|ccc}
\hline
Methods   &T-mAP &T-AP$@0.10$ &T-AP$@0.15$ &T-AP$@0.20$ \\
\hline
MCNN \cite{DBLP:conf/cvpr/ZhangZCGM16}  &$9.16/8.96$  &$11.47/10.45$ &$9.65/9.91$ &$6.36/6.51$ \\
CAN \cite{DBLP:conf/cvpr/LiuSF19} &$4.39/4.13$  &$6.97/5.48$ &$4.72/5.26$ &$1.48/1.65$ \\
CSRNet \cite{DBLP:conf/cvpr/LiZC18}   &$12.15/11.66$  &$17.34/14.63$ &$12.85/13.74$ &$6.26/6.16$ \\
DM-Count \cite{DBLP:conf/nips/WangLSH20}  &$17.01/16.54$  &$22.38/19.72$ &$18.34/19.13$ &$10.29/10.77$ \\
\hline\hline
STNNet (w/o loc) &$28.72/28.55$  &$32.52/32.50$ &$30.84/30.65$ &$22.80/22.51$ \\
STNNet (w/o ass) &$31.44/30.90$  &$34.59/34.08$ &$32.94/32.32$ &$26.77/26.30$ \\
STNNet (w/o rel) &$32.26/\bf{31.60}$  &$35.20/34.78$ &$33.78/\bf{33.12}$ &$27.80/26.89$ \\
STNNet (w/o cyc) &${\bf 32.50}/31.44$  &${\bf 35.45}/34.53$ &${\bf 33.99}/32.79$ &${\bf 28.05}/\bf{26.99}$ \\
STNNet  &$32.32/31.58$  &$35.29/\bf{34.82}$ &$33.78/33.00$ &$27.90/26.92$ \\
\hline
\end{tabular}}
\label{tab:crowd-tracking-results}
\end{minipage}
\end{table*}

{\noindent \textbf{Density map estimation.}}
As shown in Table \ref{tab:density-map-estimation-results}, our STNNet performs favorably against the state-of-the-art methods, with an improvement of $2.6$ $\text{MAE}$ and $8.3$ $\text{MSE}$ in comparison to the second best DM-Count \cite{DBLP:conf/nips/WangLSH20} in the overall {\tt testing} set. It indicates that our method generates more accurate and robust density maps in different scenarios. To further analyze the results, we report the performance on several subsets based on the video-level attributes (see Section \ref{sec:dataset}). LCFCN \cite{DBLP:conf/eccv/LaradjiRPVS18} and AMDCN \cite{DBLP:conf/cvpr/DebV18} perform not well in the {\em Crowd} subset, producing the two worst $\text{MAE}$ and $\text{MSE}$ scores. This is maybe because LCFCN \cite{DBLP:conf/eccv/LaradjiRPVS18} uses a loss function to encourage the network to output a segmentation blob for each object in crowd counting. However, in drone-captured scenarios, each object may contain only few pixels, making it difficult to separate objects accurately. AMDCN \cite{DBLP:conf/cvpr/DebV18} uses multiple columns of large dilation convolution operations, which inevitably integrates considerable background noise, affecting the accuracy in density map estimation. In contrast, MCNN \cite{DBLP:conf/cvpr/ZhangZCGM16} uses multi-column CNNs to learn the features adaptive to variations in object size due to perspective effect or image resolution, resulting in better performance. CAN \cite{DBLP:conf/cvpr/LiuSF19} achieves the best performance in both {\em Cloudy} and {\em Crowded} subsets by exploiting multi-scale contextual information in density maps. DM-Count \cite{DBLP:conf/nips/WangLSH20} obtains the best MAE and MSE scores in the {\em Sunny} subset without imposing Gaussians to annotations. Our STNNet achieves the best result in other four subsets, which demonstrates the effectiveness and importance of exploiting multi-scale features in density map estimation.

Furthermore, to study the effectiveness of the localization subnet in STNNet for density map estimation, we construct a variant of STNNet, \ie, STNNet (w/o loc), which removes the localization subnet from STNNet. As shown in Table \ref{tab:density-map-estimation-results}, our STNNet achieves better results than STNNet (w/o loc) by decreasing $2.8$ $\text{MAE}$ score and $3.5$ $\text{MSE}$ score, which validates the importance of the localization subnet. 

{\noindent \textbf{Crowd localization.}}
As presented in Table \ref{tab:crowd-localization-results}, we compare the localization results of $4$ methods with top density estimation results (\ie, MCNN \cite{DBLP:conf/cvpr/ZhangZCGM16}, CSRNet \cite{DBLP:conf/cvpr/LiZC18}, CAN \cite{DBLP:conf/cvpr/LiuSF19}, and DM-Count \cite{DBLP:conf/nips/WangLSH20}) and our STNNet variants, \ie, STNNet (w/o loc), STNNet (w/o ass) and STNNet (w/o cyc). STNNet (w/o loc) denotes the method that removes both the association and localization subnets from STNNet, STNNet (w/o ass) denotes the method that removes the association subnet from STNNet, and STNNet (w/o cyc) denotes the method that only considers the forward motion offsets in neighboring context loss computation. Meanwhile, for the density map estimation based methods such as MCNN, CSRNet, CAN, DM-Count, and STNNet (w/o loc), similar to \cite{DBLP:conf/eccv/IdreesTAZARS18}, we post-process the predicted density maps to find local peaks using a preset threshold. 

As shown in Table \ref{tab:crowd-localization-results}, we find that STNNet achieves the best accuracy with $40.45\%$ L-mAP and surpasses the second best DM-Count \cite{DBLP:conf/nips/WangLSH20} $22.28\%$ L-mAP. It indicates that our method can generate more accurate localizations of each target. Compared to STNNet (w/o cyc), STNNet improves the localization accuracy by $0.22\%$, which shows the effectiveness of cycle strategy in the neighboring context loss for the localization task.  Without the association subnet, the L-mAP score decreases $0.68\%$ ($40.45\%$ of STNNet {\em vs.} $39.77\%$), indicating that temporal coherence facilitates improve the localization accuracy. If we remove both association and localization subnets, the L-mAP score decreases more than $8\%$. It demonstrates that the localization subnet enforces the network to focus on more discriminative features to localize people's heads.

{\noindent \textbf{Crowd tracking.}}
For object tracking, two association methods, \ie, the min-cost flow method \cite{DBLP:conf/cvpr/PirsiavashRF11} and the social-LSTM method \cite{DBLP:conf/cvpr/AlahiGRRLS16}, are used to generate long trajectories of objects. To validate the effectiveness of STNNet for crowd tracking, we compare it to several methods including MCNN, CSRNet, CAN, DM-Count, STNNet (w/o loc), STNNet (w/o ass), STNNet (w/o cyc) and STNNet. It is worth mentioning that STNNet (w/o loc) performs crowd tracking based on the localized points from density maps, similar to MCNN, CSRNet, CAN, and DM-Count. Without predicting motion offsets, STNNet (w/o ass) directly associates the targets from the localization results. STNNet (w/o cyc) and STNNet first connect short tracklets in two consecutive frames based on the predicted offsets, and then generate long trajectories using the same data association methods \cite{DBLP:conf/cvpr/PirsiavashRF11,DBLP:conf/cvpr/AlahiGRRLS16}.

From Table \ref{tab:crowd-tracking-results}, we notice that STNNet achieves $32.50\%$ T-mAP score, which is $15.49\%$ higher than the second best DM-Count. Meanwhile, STNNet (w/o cyc) produces $0.11\%$ higher T-mAP score than our method. STNNet (w/o ass) produces inferior results than STNNet, \ie, $31.44\%$ vs. $32.32\%$. The T-mAP score of STNNet (w/o loc) decreases $3.60\%$ compared to STNNet (w/o ass). These results indicate that association and localization subnets are critical in crowd tracking. However, these results are still far from satisfactory. Besides, we find that the method using social-LSTM \cite{DBLP:conf/cvpr/AlahiGRRLS16} performs comparably with that using min-cost flow \cite{DBLP:conf/cvpr/PirsiavashRF11}. It indicates that it is possible to predict the motion patterns of objects based on the observed trajectories. In summary, our DroneCrowd dataset is extremely challenging for crowd tracking and much effort is needed to develop more effective methods in real scenarios.

{\noindent \textbf{Effectiveness of neighboring context loss.}}
To further demonstrate the effectiveness of the relation constraint in the neighboring context loss, we construct a variant STNNet (w/o rel) by removing the relation constraint in STNNet (w/o cyc). As shown in Table \ref{tab:crowd-localization-results} and \ref{tab:crowd-tracking-results}, STNNet (w/o rel) produces $40.00\%$ and $32.26\%$ L-mAP and T-mAP scores, respectively. STNNet (w/o cyc) improves $0.23\%$ and $0.24\%$ L-mAP and T-mAP scores compared with STNNet (w/o rel).

\section{Conclusion}
In this work, we propose the STNNet method to jointly solve density map estimation, localization, and tracking in drone-captured crowded scenes. Notably, we design the neighboring context loss to capture relations among neighboring targets in consecutive frames, which is effective for localization and tracking. To better evaluate the performances on drones, we collect and annotate a new dataset, DroneCrowd. To the best of our knowledge, it is the largest dataset to date in terms of annotated trajectories of heads for density map estimation, crowd localization, and tracking on drones. We hope the dataset and the proposed method can facilitate the research and development in crowd localization, tracking and counting on drones.

{\small
\bibliographystyle{ieee_fullname}
\bibliography{references}
}

\end{document}